\documentclass[10pt]{article}
\usepackage[margin=0.9in]{geometry}
\usepackage{amsmath,amssymb,amsthm}
\usepackage{booktabs}

\usepackage{microtype}
\usepackage{graphicx}
\usepackage{xcolor}

\usepackage[numbers,sort&compress]{natbib}
\usepackage[hidelinks]{hyperref}
\theoremstyle{plain}
\newtheorem{proposition}{Proposition}
\newcommand{\JS}{D_{\mathrm{JS}}}

\usepackage[T1]{fontenc}
\usepackage[utf8]{inputenc}

\title{Disagreeing for All the Wrong Reasons}
\author{Carlos Bernardino\textsuperscript{1}\and Joao Tiago Aparicio\textsuperscript{2}}
\date{\small\textsuperscript{1}Portuguese Air Force Academy, Centro de Investigação da Força Aérea, Instituto Universitário Militar, Sintra, Portugal\\
\small\textsuperscript{2}INESC-ID, Instituto Superior T\'ecnico, Universidade de Lisboa, Lisbon, Portugal\\[6pt]
November 2026}
\begin{document}
\maketitle
\begin{abstract}
Reviewers can inspect only a small fraction of a classifier's decisions, so a useful uncertainty score must bring its errors to the front of the queue. An ensemble offers two signals for this task, the entropy of its mean prediction and disagreement among its members. We compare them with the classifier's labels held fixed. In a two-regime Gaussian logit model, member scatter can help identify errors when its variance differs between regimes. The main result bounds the change in conditional error probability attributable to scatter by a quantity proportional to the inverse of ensemble size. More members can therefore make the regime easier to distinguish while reducing the extra error information it provides. This result concerns classification by the mean logit. Separately, simulations using mean probabilities cover ensembles of 3 to 50 members, five scatter ratios and 200 repeated studies per setting. At a 10\% review budget, disagreement alone captures 13 to 65 percentage points fewer errors than entropy. A score derived from the mean-logit posterior gains at most 3.7 points; it is a comparator, not an optimal bound for the simulated classifier. These findings explain one reason why a clearer disagreement signal need not produce a better review queue. %They apply to the stated model and do not establish performance for trained detectors or human reviewers.
\end{abstract}
\section{Introduction}
Synthetic video can impersonate people, invent events and lend weight to misleading stories. Newsrooms, platforms and open-source investigators who meet this material run automated detectors first and send part of the output to people for checking. That part is small. Even a detector with 5\% errors leaves open which 5\%, since reviewers can open only a limited number of clips and the clips worth their time are those where the automated label is wrong. A detector score says nothing about whether the claim attached to a video is true; it only helps decide where human attention goes first.

Improving accuracy and improving a review queue are different goals. A change to the decision rule can raise accuracy and still make the queue less useful, and a ranking score can expose many errors without altering a single label. We therefore freeze the labels of an ensemble and vary only the score that orders the queue.

Any ensemble supplies two ordering signals. One is the uncertainty of the averaged prediction, usually summarised by entropy. The other is the disagreement among members, measured by the Jensen--Shannon (JS) divergence between their predicted distributions. Here we ask when disagreement adds useful error information beyond mean confidence, and how that opportunity changes as members are added.

Our main result is Proposition~\ref{prop:prize}. In a two-regime Gaussian model, the effect of member scatter on conditional error probability is bounded by a quantity proportional to $1/M$, where $M$ is ensemble size. At the same time, the distributions of scatter in the two regimes become easier to distinguish. These are different statements. Knowing more about the regime need not reveal much more about whether the ensemble is wrong. The simulations examine this distinction for a fixed mean-probability classifier; the proofs concern a mean-logit classifier, and we keep that boundary explicit.

\section{Related work}

\subsection{Ensembles, confidence and disagreement}

Deep ensembles give useful predictive variability without a Bayesian model \cite{lakshminarayanan2017}. For members with positive-class probabilities $p_m$ and mean $\bar p=\frac1M\sum_m p_m$, generalised JS disagreement \cite{lin1991} is
\begin{equation}
\JS(x)=H(\bar p(x))-\frac1M\sum_{m=1}^{M}H(p_m(x)),
\label{eq:js}
\end{equation}
where $H$ is binary entropy in nats, with natural logarithms. Rearranged, the entropy of the mean splits into two parts,
\begin{equation}
H(\bar p)=A+\JS,\qquad A=\frac1M\sum_{m=1}^{M}H(p_m),
\label{eq:split}
\end{equation}
where $A$ is the average doubt inside each member and $\JS$ is the doubt created by members disagreeing with one another. The comparison in this study asks whether $\JS$ ranks errors better than the sum $A+\JS$ that contains it. Independently trained architectures are not posterior samples, so we call $\JS$ between-detector disagreement and make no claim that it is a calibrated probability of error. Temperature scaling can fix source-domain calibration, but calibration can degrade under shift \cite{guo2017,ovadia2019}.

\subsection{Selective prediction and review allocation}

Selective classification trades coverage against risk \cite{geifman2017}. Risk--coverage curves summarise that trade across all budgets. A review team works at one budget. Here we measure the share of the system's errors that enter the queue, with error-ranking AUROC as a secondary summary in the accompanying results.

\subsection{Deepfake detection as motivation}

Wahab et al.\ average six deepfake detectors \cite{wahab2025ensemble}. Farooq et al.\ learn disagreement-aware uncertainty inside a multi-branch network \cite{farooq2026discern}, and Correlation-Optimized Fusion combines several uncertainty sources \cite{sharma2026cof}. Collins, Li and Nguyen compare fusion and per-video routing across 22 datasets and find that apparent routing gains vanish under family-level holdouts \cite{collins2026honest}. Their routing changes the fused score and can change the predicted label. Our comparison freezes one classifier and asks how ranking scores order its errors. This question also follows the authors' earlier work on abstention-aware deepfake screening \cite{bernardino2026deepfake}. No trained detector is evaluated here.

\section{Materials and methods}

\subsection{A two-regime model}

Each synthetic case has a label $y\in\{-1,+1\}$ with equal class probabilities and belongs to an ordinary regime ($z=0$) or a shifted regime ($z=1$, probability $\pi_1$). Member $m$ returns the logit $a_m=y\mu+u+\varepsilon_m$, where $\mu>0$ is the class margin, $u\sim N(0,s^2)$ with $s>0$ is ambiguity shared by all members, and $\varepsilon_m\mid z\sim N(0,\tau_z^2)$ is member-specific noise. We assume $M\ge2$, $0<\pi_1<1$ and $0<\tau_0\le\tau_1$. Labels and regimes are independent; $u$ is independent of both, and the member noises are independent conditional on $z$ and independent of $y$ and $u$. Cases are drawn independently.

Write $\bar a=M^{-1}\sum_m a_m$, $r=|\bar a|$, and $S^2=(M-1)^{-1}\sum_m(a_m-\bar a)^2$. The within-case variance with divisor $M$ is $\hat\sigma_a^2=(M-1)S^2/M$. This distinction matters in the small-scatter approximation below.

\subsection{Two aggregation rules}

The theory uses $\hat y_L=\operatorname{sign}(\bar a)$, with a positive label at zero. The simulations instead use $p_m=\sigma(a_m)$, where $\sigma(t)=(1+e^{-t})^{-1}$, and freeze the mean-probability label
\begin{equation}
\bar p=\frac1M\sum_{m=1}^{M}p_m,\qquad
\hat y_P=\begin{cases}+1,&\bar p\ge0.5,\\-1,&\bar p<0.5.\end{cases}
\label{eq:prediction}
\end{equation}
These rules are not interchangeable. For example, logits $(3,3,-7)$ have negative mean but a mean probability above $0.5$. For small logit scatter, $\bar p-\sigma(\bar a)$ is of second order, but this does not guarantee equal labels near the decision boundary. All simulated errors below are errors of $\hat y_P$.

\subsection{Rankings and review queues}

The baseline ranks by $H(\bar p)$ and the primary alternative by $\JS$. The accompanying results also include sample logit variance $S^2$, mean member entropy $A$, random order, and a fitted score using entropy and log dispersion. The fitted score is a logistic regression of the mean-probability classifier's error indicator on $H(\bar p)$ and $\log(S^2+10^{-9})$. Features are standardised on one independent validation draw of 15 groups of 1,006 cases per parameter setting. The code uses 25 Newton steps with a quadratic penalty of $10^{-3}\|w\|^2/2$, including the intercept, and holds this fit fixed across test replicates.

We also evaluate the posterior error score derived for $\hat y_L$ in Proposition~\ref{prop:regime}. We call it the \emph{logit-posterior score}. Applied to errors of $\hat y_P$, it is a comparator and has no optimality guarantee. In the result files it retains the original key \texttt{posterior}.

At budget $b$, the queue $Q_b$ holds the top $\lceil bN\rceil$ cases in each group. Scores are sorted in descending order, with seeded pseudorandom tie-breaking as implemented in the supplied code. Every ranking uses the same cases, labels and queue size. Let $E=\{i\mathrel{|}\hat y_{P,i}\ne y_i\}$ be the set of errors. The primary outcome is
\begin{equation}
\mathrm{ErrorCapture@}b=\frac{|Q_b\cap E|}{|E|}.
\label{eq:capture}
\end{equation}
Capture is undefined for a group with no errors. The implementation omits such groups from each replicate's equal-weight mean; the omission is identical for all rankings because the labels are fixed. A replicate with no eligible groups would be undefined. Error-ranking AUROC likewise omits groups without both correct and incorrect labels. This conditional averaging rule is particularly relevant in the low-error sensitivity settings.

\subsection{Simulation design and reproducibility}

The main grid contains 40 settings, with $M\in\{3,5,7,10,15,20,30,50\}$ and $\tau_1/\tau_0\in\{1,2,3,4,6\}$. Parameters are $\mu=1$, $s=0.5$, $\tau_0=0.3$ and $\pi_1=0.4$. Each setting has 200 independent test replicates, each containing 15 groups of 1,006 independent synthetic cases. The group sizes were motivated by a possible video benchmark evaluation, but the simulation contains no matched real/fake pairs, source clusters or group-specific parameter differences. Classes are balanced in expectation, not forced to be equal in each group.

We report capture at a 10\% budget and curves from 1\% to 50\%. Paired differences are computed within each test replicate before averaging. The intervals shown are the 2.5th and 97.5th percentiles across replicates. They describe variation between simulated studies, conditional on one validation fit; they are not confidence intervals for the mean gain. No multiplicity-adjusted significance test is claimed. The sensitivity analysis uses four parameter settings, including the main setting, at ratio 3 and $M\in\{3,10,30\}$, with 50 test replicates per setting.

The code uses NumPy's \texttt{default\_rng}. Main-grid seeds are $1000+i$ for zero-based cells ordered by increasing $M$, then increasing scatter ratio. Sensitivity seeds are $5000+10j+M$, where $j=0,1,2,3$ follows the parameter order $(1,0.5,0.3)$, $(1,0.8,0.5)$, $(1.5,0.5,0.3)$ and $(0.8,0.4,0.2)$ for $(\mu,s,\tau_0)$. Numerical checks use seed 7. The accompanying material contains the executable simulation and plotting scripts, the four result tables and the dependency versions. No deepfake dataset, detector checkpoint or human-participant data were used.

\section{Results}
\label{sec:theory}

\subsection{When can scatter add error information?}

The identity in Eq.~\eqref{eq:split} says that disagreement is one additive part of the entropy that the baseline already uses. The next four propositions describe when scatter adds information under the stated Gaussian model. In Propositions~\ref{prop:regime} and~\ref{prop:prize}, error means $\hat y_L\ne y$.

\begin{proposition}[Disagreement at small logit scatter]\label{prop:scatter}
Let $a_m=\bar a+t b_m$ for fixed $\bar a$ and $b_m$ satisfying $\sum_m b_m=0$. As $t\to0$,
\[
\JS=\tfrac12\bar p(1-\bar p)\hat\sigma_a^2+O(t^3),
\qquad \hat\sigma_a^2=\frac1M\sum_m(a_m-\bar a)^2.
\]
\end{proposition}

To second order, the disagreement score combines mean uncertainty with logit scatter. Since $\bar p(1-\bar p)$ increases with $H(\bar p)$, equal scatter gives the same ordering at this order. Higher-order terms can change that ordering when scatter is appreciable.

\begin{proposition}[Scatter and conditional error]\label{prop:regime}
Let $v_z=s^2+\tau_z^2/M$ and $q_z(r)=(1+e^{2\mu r/v_z})^{-1}$. Then
\[
\eta(r,S^2)=P(\hat y_L\ne y\mid r,S^2)
=(1-w)q_0(r)+wq_1(r),
\]
where $w=P(z=1\mid r,S^2)$.
(i) If $\tau_1>\tau_0$, $\eta$ is strictly increasing in $S^2$ at every fixed $r>0$.
(ii) If $\tau_1=\tau_0$, it depends on $r$ alone, and entropy of the mean-logit probability, $H(\sigma(\bar a))$, ranks errors optimally in expected count.
(iii) For independent cases and a fixed queue size, ranking by $\eta$ maximises the conditional expected number of errors of $\hat y_L$ captured by the queue.
\end{proposition}

\begin{proposition}[More members distinguish the regimes]\label{prop:info}
Conditional on $z$, $S^2\sim\tau_z^2\chi^2_{M-1}/(M-1)$. If $g_z$ is its density in regime $z$ and $\theta=\tau_1^2/\tau_0^2$, then
\[
D_{\mathrm{KL}}(g_1\|g_0)=\frac{M-1}{2}(\theta-1-\ln\theta).
\]
For fixed $\theta>1$, this directed divergence grows linearly in $M-1$.
\end{proposition}

This is a divergence between two conditional distributions, not the mutual information between $z$ and $S^2$. Mutual information is bounded by the binary regime entropy, $I(z;S^2)\le H(z)\le\ln2$. A growing KL divergence does not mean unbounded information about a binary regime.

\begin{proposition}[More members shrink the extra error signal]\label{prop:prize}
Under the model, for every $r\ge0$,
\[
0\le q_1(r)-q_0(r)
\le \frac{0.23(\tau_1^2-\tau_0^2)}{M s^2}.
\]
Both $\eta(r,S^2)$ and $P(\hat y_L\ne y\mid r)$ lie between $q_0(r)$ and $q_1(r)$. Their absolute difference is therefore bounded by the same quantity, which is $O(M^{-1})$ when the model parameters are held fixed.
\end{proposition}

Proposition~\ref{prop:prize} is the central result. With three members, $S^2$ has only two degrees of freedom. Adding members makes the regimes easier to distinguish, but reduces the change in error probability that knowing the regime can explain. The bound is pointwise in $r$. It does not by itself bound the difference in ErrorCapture, prove that capture gains decrease monotonically with $M$, or apply to arbitrary trained ensembles. Proofs are in the Appendix. Fig.~\ref{fig:checks} checks the small-scatter approximation and directed KL formula numerically.

\begin{figure}[!htbp]
\centering
\includegraphics[width=\linewidth]{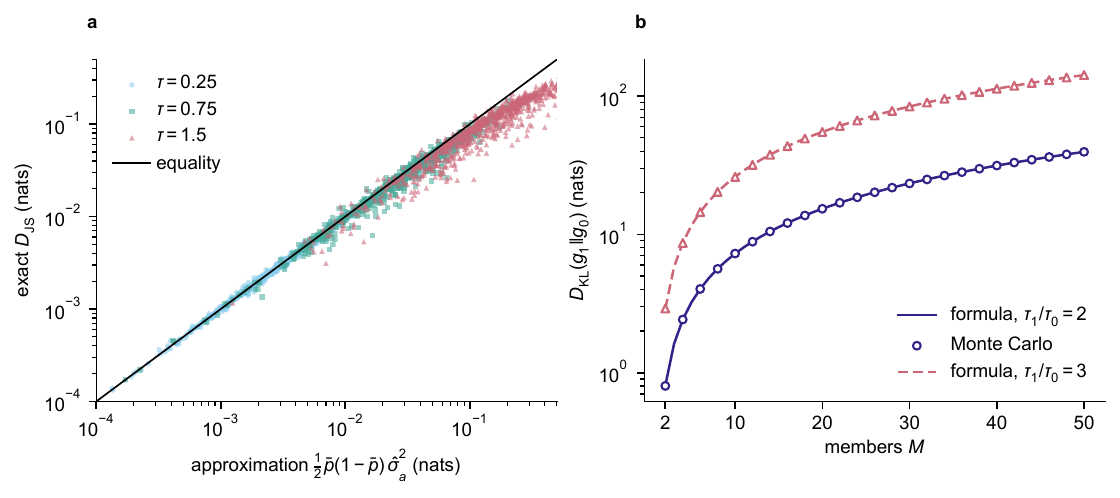}
\caption{Numerical checks of the approximation and regime divergence. \textbf{a}, Exact JS disagreement against the second-order expression for 800 synthetic cases at each of three scatter levels, with $M=5$. The approximation becomes less accurate as scatter grows. \textbf{b}, Directed $D_{\mathrm{KL}}(g_1\|g_0)$ in nats from Proposition~\ref{prop:info} (lines) and $4\times10^5$ Monte Carlo draws per setting (open markers), for 2 to 50 members. This is not mutual information or a classification accuracy. Plot labels and layout were revised with OpenAI Codex assistance; the plotted data were retained.}
\label{fig:checks}
\end{figure}

\subsection{Simulation results for mean-probability labels}

\begin{figure}[!htbp]
\centering
\includegraphics[width=\linewidth]{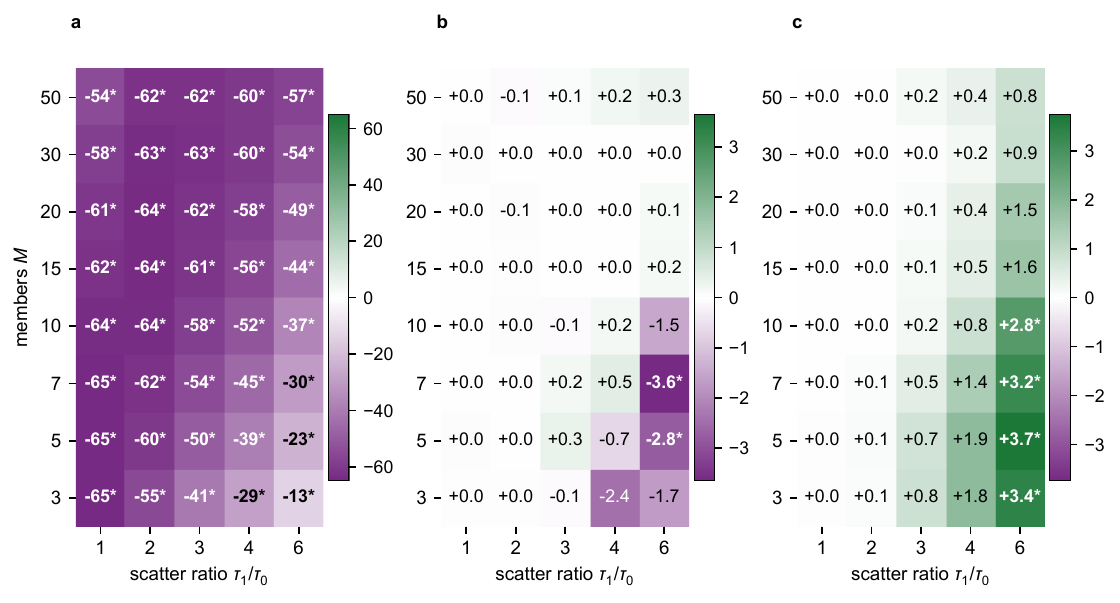}
\caption{Change in error capture at a 10\% review budget relative to entropy of the mean probability. Values are percentage-point differences averaged over 200 test replicates per cell. \textbf{a}, JS disagreement. \textbf{b}, Entropy-plus-dispersion score fitted on one independent validation draw. \textbf{c}, Logit-posterior score evaluated on mean-probability classification errors, without an optimality guarantee. Asterisks mark cells whose central 95\% replicate interval excludes zero; they do not denote a significance test. All colour scales are centred at zero, with different ranges across panels. Plot labels and layout were revised with OpenAI Codex assistance; the plotted data were retained.}
\label{fig:heat}
\end{figure}

Ensemble error rates in the main grid ranged from 2.3\% to 9.7\%. Disagreement alone captured fewer errors than entropy in all 40 cells, by 13 to 65 percentage points (Fig.~\ref{fig:heat}a). The loss was smallest where shift scatter was largest. The small-scatter approximation helps explain this result, though it is less accurate at large scatter. Disagreement combines proximity to the decision boundary with member scatter; scatter can raise the score of cases whose mean prediction remains confident.

\begin{figure}[!htbp]
\centering
\includegraphics[width=\linewidth]{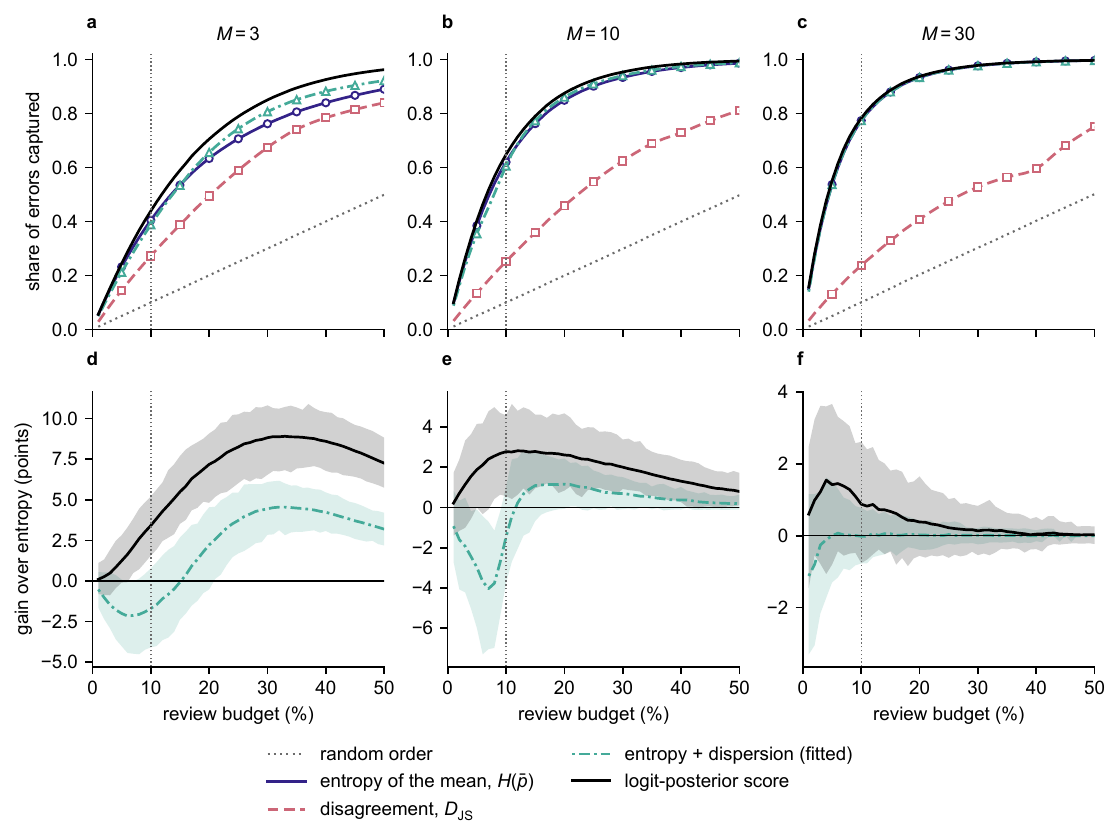}
\caption{Error capture across review budgets at $\tau_1/\tau_0=6$. \textbf{a--c}, Mean capture over 200 test replicates for 3, 10 and 30 members. \textbf{d--f}, Paired gain over mean-probability entropy for the fitted score and the logit-posterior score. Shading gives central 95\% replicate intervals, conditional on the validation fit. The dotted vertical line marks the 10\% budget. Plot labels and layout were revised with OpenAI Codex assistance; the plotted data were retained.}
\label{fig:curves}
\end{figure}

The logit-posterior score gained at most 3.7 points, with a replicate interval wholly above zero only at $\tau_1/\tau_0=6$ and $M\le10$ (Fig.~\ref{fig:heat}c). The fitted two-feature score gained little and had negative mean gains in several cells, with replicate intervals wholly below zero in two (Fig.~\ref{fig:heat}b). At $\tau_1/\tau_0=6$, the logit-posterior gain fell from 3.4 points at $M=3$ to 0.8 points at $M=50$, while the loss of JS alone grew from 13 to 57 points. This pattern is consistent with the mechanism in Proposition~\ref{prop:prize}, but is a separate numerical observation because the simulated label rule differs. Fig.~\ref{fig:curves} shows the full curves at $\tau_1/\tau_0=6$ for $M=3$, 10 and 30.

Table~\ref{tab:sens} repeats the comparison at scatter ratio 3 under four parameter settings, including the main setting. Disagreement loses in all twelve rows and the logit-posterior score gains at most 0.8 points. Each row uses 50 test replicates. Groups with no errors are omitted from capture averages as specified in the methods.

\begin{table}[!htbp]
\centering
\small
\caption{Sensitivity of ErrorCapture@10\% at scatter ratio 3. Differences relative to entropy are in percentage points. Fitted denotes the validation-fitted entropy-plus-dispersion score; logit-post. denotes the logit-posterior comparator.}
\label{tab:sens}
\begin{tabular}{@{}cccrrrrr@{}}
\toprule
$\mu$ & $s$ & $\tau_0$ & $M$ & errors (\%) & $D_{\rm JS}$ & fitted & logit-post. \\
\midrule
1 & 0.5 & 0.3 & 3 & 5.1 & -40.7 & +0.2 & +0.8 \\
1 & 0.5 & 0.3 & 10 & 3.1 & -57.8 & +0.0 & +0.2 \\
1 & 0.5 & 0.3 & 30 & 2.5 & -62.2 & +0.0 & +0.0 \\
1 & 0.8 & 0.5 & 3 & 15.1 & -11.8 & -1.2 & +0.3 \\
1 & 0.8 & 0.5 & 10 & 12.3 & -18.7 & +0.0 & +0.4 \\
1 & 0.8 & 0.5 & 30 & 11.1 & -20.7 & -0.5 & +0.3 \\
1.5 & 0.5 & 0.3 & 3 & 0.9 & -59.4 & +0.0 & +0.7 \\
1.5 & 0.5 & 0.3 & 10 & 0.3 & -61.5 & +0.0 & +0.1 \\
1.5 & 0.5 & 0.3 & 30 & 0.2 & -63.4 & +0.0 & +0.0 \\
0.8 & 0.4 & 0.2 & 3 & 4.2 & -49.7 & -0.3 & +0.7 \\
0.8 & 0.4 & 0.2 & 10 & 2.9 & -63.2 & -0.1 & +0.1 \\
0.8 & 0.4 & 0.2 & 30 & 2.5 & -65.7 & +0.0 & +0.0 \\
\bottomrule
\end{tabular}

\end{table}

\section{Discussion}

A review team needs to know which cases to open first. Improving the classifier and improving the queue are different goals. Keeping the classifier's labels fixed makes that distinction visible, because any change in captured errors comes from the ranking alone.

The analysis narrows the question. Entropy already contains disagreement as one additive part. Under the Gaussian model, scatter adds conditional error information when its variance differs between regimes. Proposition~\ref{prop:prize} shows that the extra signal shrinks with ensemble size even as the regime distributions become more distinguishable. A clearer estimate of disagreement is therefore not, on its own, a reason to expect a better queue.

The simulations give a concrete example. Disagreement alone loses substantially to entropy, and the fitted combination adds little in the settings examined. The logit-posterior comparator does better in some cells, but its theoretical optimality concerns a different label rule. Its observed gains cannot be presented as the most any ranking could achieve for the simulated classifier.

\section{Conclusions and future work}

More ensemble members can make a high-scatter regime easier to recognise while leaving less additional error information for scatter to supply. Proposition~\ref{prop:prize} makes this trade-off precise for mean-logit classification in the stated Gaussian model. In the accompanying mean-probability simulations, disagreement alone captures fewer errors than entropy at a fixed review budget. These results support checking what disagreement adds beyond mean confidence, rather than assuming that a more stable disagreement estimate must improve review allocation.

The model assumes equal class priors, Gaussian ambiguity, and member noises that are independent conditional on the regime. Members still share the ambiguity term $u$. Real ensembles can have unequal calibration, systematic shared errors, non-Gaussian tails and dependence not represented by this construction. The $1/M$ bound holds for fixed $s>0$ and fixed regime variances. It does not establish a general law for deepfake detectors.

The simulation groups contain independent cases, not repeated descendants of shared sources. The replicate intervals also condition on one fitted validation model per setting. They do not quantify uncertainty over validation fits or training runs. Error capture counts which classifier errors enter a queue; it says nothing about whether reviewers correct them or how much time they save.

A future evaluation could compare entropy and disagreement for fixed labels from GenD, LAA-Net and FakeSTormer on SynthForensics \cite{leotta2026synthforensics}. That study would need verified source-grouped splits, training-overlap checks and a validation-only choice of calibration and analysis rules. It is not part of the evidence reported here. A separate study with reviewers would be needed to assess practical benefit.

%\section*{Data and code availability}
%The simulation code, generated result tables and reproduction instructions accompany this manuscript as ancillary material.

\section*{Competing interests}
The authors declare no competing interests.

%\section*{Acknowledgments}
%This work builds on the authors' earlier research on the Atomic Evaluation Framework. OpenAI Codex was used to assist with language editing, mathematical consistency checks, figure labels and layout, code documentation, reproducibility checks and submission formatting. Simulation values were retained from the supplied result files. The preparation checks included rerunning the supplied simulation, comparing numerical outputs, compiling the manuscript and inspecting the rendered pages.

\bibliographystyle{unsrtnat}
\bibliography{refs}

\clearpage
\appendix
\section{Proofs}

\textbf{Proposition~\ref{prop:scatter}.} Write $p_m=\bar p+\delta_m$, with $\sum_m\delta_m=0$. A second-order expansion gives
\[
H(p_m)=H(\bar p)+H'(\bar p)\delta_m
+\tfrac12H''(\bar p)\delta_m^2+O(t^3).
\]
Averaging removes the linear term. Since $H''(p)=-1/[p(1-p)]$ for entropy in nats,
\[
\JS=\frac{\overline{\delta^2}}{2\bar p(1-\bar p)}+O(t^3).
\]
Smoothness of $\sigma$ at fixed $\bar a$ and $\sum_m b_m=0$ give $\bar p=\sigma(\bar a)+O(t^2)$ and $\delta_m=\bar p(1-\bar p)t b_m+O(t^2)$. Substitution yields the claim with divisor $M$.

\textbf{Proposition~\ref{prop:regime}.} Conditional on $y,z$, $\bar a\sim N(y\mu,v_z)$. The deviations $a_m-\bar a=\varepsilon_m-\bar\varepsilon$ do not involve $y$ or $u$ and are independent of $\bar\varepsilon$ for Gaussian noise. Thus $S^2$ and $\bar a$ are independent given $z$. With equal label priors, the ratio of the densities at $-r$ and $+r$ for the positive class is $e^{-2\mu r/v_z}$. Symmetry gives $P(\hat y_L\ne y\mid r,z)=q_z(r)$.

Let $f_z$ and $g_z$ be the densities of $r$ and $S^2$ given $z$. The posterior log-odds of $z=1$ are
\[
\operatorname{logit}(\pi_1)+\ln\frac{f_1(r)}{f_0(r)}
+\ln\frac{g_1(S^2)}{g_0(S^2)}.
\]
The last term equals a constant minus $\tfrac{M-1}{2}S^2(\tau_1^{-2}-\tau_0^{-2})$, which is strictly increasing in $S^2$ when $\tau_1>\tau_0$. Then $v_1>v_0$ and $q_1(r)>q_0(r)$ for $r>0$, proving (i). If $\tau_1=\tau_0$, $q_0=q_1$ and this error probability decreases with $r$, just as $H(\sigma(\bar a))$ does, proving (ii).

For (iii), the Gaussian residual density depends on the deviations only through $S^2$. Their remaining direction is ancillary for $y$ and $z$, so the same posterior error probability is obtained from the full logit vector. For independent cases, the conditional expected number of errors in a fixed-size queue is the sum of its individual posterior error probabilities. The largest such sum is obtained by choosing the largest probabilities. This argument concerns error counts for $\hat y_L$, not an optimality claim for the mean-probability error-capture curves.

\textbf{Proposition~\ref{prop:info}.} $(M-1)S^2/\tau_z^2\sim\chi^2_{M-1}$, so $g_z$ is a Gamma density with common shape $\alpha=(M-1)/2$ and scale $2\tau_z^2/(M-1)$. The expected log density ratio under $g_1$ is $D_{\mathrm{KL}}(g_1\|g_0)=\alpha(\theta-1-\ln\theta)$.

\textbf{Proposition~\ref{prop:prize}.} Put $x=2\mu r/v$ and $q(v)=(1+e^x)^{-1}$. Then
\[
\frac{\partial q}{\partial v}=\frac{1}{v}\frac{x e^x}{(1+e^x)^2}.
\]
For $x\ge0$, $\cosh(x/2)\ge1+x^2/8$, so
\[
\frac{x}{4\cosh^2(x/2)}
\le\frac{x}{4(1+x^2/8)^2}
\le\frac{3\sqrt6}{32}<0.23.
\]
The middle expression is maximised at $x=\sqrt{8/3}$. Hence
\[
q_1(r)-q_0(r)
\le0.23\int_{v_0}^{v_1}\frac{dv}{v}
\le0.23\frac{v_1-v_0}{v_0}
\le\frac{0.23(\tau_1^2-\tau_0^2)}{Ms^2}.
\]
Both conditional error probabilities are convex combinations of $q_0(r)$ and $q_1(r)$, which gives the stated difference bound.

\end{document}